\documentclass[runningheads]{llncs}
\usepackage{perpage} 
\MakePerPage{footnote} 
\usepackage{graphicx}

\usepackage{color,soul}
\usepackage{dsfont}
\usepackage{amsmath}
\DeclareMathOperator*{\argmax}{\arg\!\max}
\DeclareMathOperator*{\argmin}{\arg\!\min}

\begin{document}

\title{Deep Information Theoretic Registration}
%
%

\author{}
\author{Alireza Sedghi \inst{1}
\and
Jie Luo \inst{2,4}
\and
Alireza Mehrtash \inst{2,5}
\and
Steve Pieper \inst{2}
\and
Clare M. Tempany \inst{2}
\and
Tina Kapur \inst{2}
\and
Parvin Mousavi \inst{1}
\and
William M. Wells III \inst{2,3}
}
\authorrunning{Sedghi et al.}
%

\institute{Medical Informatics Laboratory, Queen's University, Kingston, Canada\\
\and
Radiology Department, Brigham and Women’s Hospital, Harvard Medical School, Boston, USA
\and
Computer Science and Artificial Intelligence Laboratory, Massachusetts Institute of Technology, Cambridge, USA
\and
Graduate School of Frontier Sciences, The University of Tokyo, Japan
\and
Department of Electrical and Computer Engineering, University of British Columbia, Vancouver, Canada
}

\maketitle              
\begin{abstract}
This paper establishes an information theoretic framework for deep metric based image registration techniques. We show an exact equivalence between maximum profile likelihood and minimization of joint entropy, an important early information theoretic registration method. We further derive deep classifier-based metrics that can be used with iterated maximum likelihood to achieve Deep Information Theoretic Registration on patches rather than pixels. This alleviates a major shortcoming of previous information theoretic registration approaches, namely the implicit pixel-wise independence assumptions. Our proposed approach does not require well-registered training data; this  brings previous fully supervised deep metric registration approaches to the realm of weak supervision. We evaluate our approach on several image registration tasks and show significantly better performance compared to mutual information, specifically when images have substantially different contrasts. This work enables general-purpose registration in applications where current methods are not successful.

\end{abstract}


\newcommand{\cf}{\S}
\newcommand\eg{\textit{e.g.~}}
\newcommand\etal{\textit{et~al.~}}
\newcommand\ie{\textit{i.e.~}}
\newcommand\viz{\textit{viz.~}}
\newcommand{\App}[1]{Appendix~\ref{#1}}
\newcommand{\Algo}[1]{Algorithm~\ref{#1}}
\newcommand{\Eqn}[1]{eqn.~\eqref{#1}}
\newcommand{\eqn}[1]{eqn.~\eqref{#1}}
\newcommand{\Chap}[1]{Chapter~\ref{#1}}
\newcommand{\Fig}[1]{Fig.~\ref{#1}}
\newcommand{\Sec}[1]{Section~\ref{#1}}
\newcommand{\Table}[1]{Table~\ref{#1}}

\newcommand\ba{\mathbf{a}}
\newcommand\bb{\mathbf{b}}
\newcommand\bB{\mathbf{B}}
\newcommand\bc{\mathbf{c}}
\newcommand\bC{\mathbf{C}}
\newcommand\bd{\mathbf{d}}
\newcommand\bD{\mathbf{D}}
\newcommand\be{\mathbf{e}}
\newcommand\bff{\mathbf{f}}
\newcommand\bg{\mathbf{g}}
\newcommand\bK{\mathbf{K}}
\newcommand\bm{\mathbf{m}}
\newcommand\bn{\mathbf{n}}
\newcommand\bp{\mathbf{p}}
\newcommand\br{\mathbf{r}}
\newcommand\bs{\mathbf{s}}
\newcommand\bt{\mathbf{t}}

\newcommand\bu{\mathbf{u}}
\newcommand\bU{\mathbf{U}}
\newcommand\bv{\mathbf{v}}
\newcommand\bw{\mathbf{w}}
\newcommand\bx{\mathbf{x}}
\newcommand\by{\mathbf{y}}
\newcommand\bz{\mathbf{z}}
\newcommand\balpha{\boldsymbol{\alpha}}
\newcommand\bbeta{\boldsymbol{\beta}}
\newcommand\bepsilon{\boldsymbol{\epsilon}}
\newcommand\blambda{{\boldsymbol{\lambda}}}
\newcommand\bLambda{\boldsymbol\Lambda}
\newcommand\bmu{\boldsymbol{\mu}}
\newcommand\bsigma{\boldsymbol{\sigma}}
\newcommand\bSigma{\boldsymbol{\Sigma}}
\newcommand\btheta{\boldsymbol{\theta}}

\newcommand\bbE{\mathbb{E}}
\newcommand\bbG{\mathbb{G}}
\newcommand\bbR{\mathbb{R}}
\newcommand\bbV{\mathbb{V}}
\newcommand\bbZ{\mathbb{Z}}

\newcommand\cA{\mathcal{A}}
\newcommand\cB{\mathcal{B}}
\newcommand\cD{\mathcal{D}}
\newcommand\cE{\mathcal{E}}
\newcommand\cF{\mathcal{F}}
\newcommand\cG{\mathcal{G}}
\newcommand\cI{\mathcal{I}}
\newcommand\cJ{\mathcal{J}}
\newcommand\cL{\mathcal{L}}
\newcommand\cM{\mathcal{M}}
\newcommand\cN{\mathcal{N}}
\newcommand\cO{\mathcal{O}}
\newcommand\cP{\mathcal{P}}
\newcommand\cT{\mathcal{T}}
\newcommand\cV{\mathcal{V}}
\newcommand\cU{\mathcal{U}}
\newcommand\cX{\mathcal{X}}
\newcommand\cZ{\mathcal{Z}}

\newcommand\dsE{\mathds{E}}
\newcommand\dsG{\mathds{G}}
\newcommand\dsH{\mathds{H}}
\newcommand\dsV{\mathds{V}}

\newcommand\rmA{\mathrm{A}}
\newcommand\rmB{\mathrm{B}}
\newcommand\rmd{\mathrm{d}}
\newcommand\rmD{\mathrm{D}}
\newcommand\rmE{\mathrm{E}}
\newcommand\rmF{\mathrm{F}}
\newcommand\rmH{\mathrm{H}}
\newcommand\rmI{\mathrm{I}}
\newcommand\rmJ{\mathrm{J}}
\newcommand\rmK{\mathrm{K}}
\newcommand\rmM{\mathrm{M}}
\newcommand\rmT{\mathrm{T}}
\newcommand\rmU{\mathrm{U}}
\newcommand\rmV{\mathrm{V}}
\newcommand\rmW{\mathrm{W}}
\newcommand\rmy{\mathrm{y}}
\newcommand\rmz{\mathrm{z}}
\newcommand\rmZ{\mathrm{Z}}


\newcommand{\distributionequal}{\operatornamewithlimits{\sim}}
\newcommand{\diag}{\textrm{diag}}
\newcommand{\const}{\textrm{const}}
\newcommand{\Cov}{\mathds{C}\text{ov}}
\newcommand{\Corr}{\mathds{C}\text{orr}}
\newcommand{\E}[1]{{\dsE\left\{#1\right\}}}
\newcommand{\Ent}[1]{{\dsH\left\{#1\right\}}}
\newcommand{\Exp}{\dsE}
\newcommand{\half}{\frac{1}{2}}
\newcommand\Hes{\nabla\nabla\tr}
\newcommand{\inv}{^{-1}}
\newcommand{\IP}[1]{{\left\langle{#1}\right\rangle}} 
\newcommand\I{\rmI}
\newcommand\ind{\mathds{1}}
\newcommand\KL{\mathds{KL}}
\newcommand\one{\mathds{1}}
\newcommand{\p}[1]{{p\left(#1\right)}}
\newcommand{\pinv}{^{-}}
\newcommand{\Prec}{\bLambda}
\newcommand{\Prr}{\mathrm{Pr}}
\newcommand\R{\bbR}
\newcommand\rank{{\mathrm{Rank}}}
\newcommand{\tr}{^\top}
\newcommand{\Trace}{\mathrm{Tr}}
\newcommand\Var[1]{{\mathds{V}\mathrm{ar}\left\{#1\right\}}}
\newcommand\vol{\mathrm{vol}}


\newcommand{\TV}{\mathrm{TV}}
\newcommand{\img}{\mathrm{y}}
\renewcommand{\ij}{{i,j}}
\newcommand{\ioj}{{i+1,j}}
\newcommand{\ijo}{{i,j+1}}
\newcommand{\iojo}{{i+1,j+1}}
\newcommand{\x}{{x}}
\newcommand{\y}{{\rmy}}
\newcommand{\z}{{\rmz}}
\newcommand{\q}{{q}}
\newcommand{\Q}{{Q}}
\newcommand{\Py}{{P}}
\renewcommand{\E}[1]{{\dsE_\Q\left\{#1\right\}}}

\newcommand{\KLD}[2]{{\mathds{KL}}[#1 || #2]}
\newcommand{\EV}[2]{{\dsE}_{#1}\left[#2\right]}
\newcommand{\PD}{~.}
\newcommand{\CM}{~,}
\newcommand{\zvar}{\rmz}
\newcommand{\area}{}
\newcommand{\wrt}{wrt.~}
\newcommand{\ase}{\overset{a.s.}{=}}
\newcommand\where{\text{where }}
\newcommand\thrfr{\text{therefore }}

\newcommand{\indx}{i\in{\mathbb I}_{\cX}}

\renewcommand{\Ent}[1]{{\dsH\left[#1\right]}}

\newcommand{\INDEX}{{\mathcal{I}}}
\newcommand{\CAT}{{\rm CAT}}
\newcommand{\TF}[2]{{^{#2}#1}}
\newcommand{\ENT}[1]{{\rm H}\left[#1\right]}









\section{Introduction}

In multi-modality image registration, mutual information (MI) and its variants
have resulted in notable successes, solving many problems ``out of the box'' 
\cite{wells1996multi,studholme1999overlap,maes1997multimodality,pluim2003mutual}. Since these methods do not require any data beyond the images being registered, they can be referred to as ``unsupervised''. Despite their strength, MI and its variants do not perform well for inter-modality image registration where, e.g.,  one modality has ``tissue contrast'' while the other has ``boundary contrast'' (e.g., CT to ultrasound registration). Here, a metric designed specifically for the application, such as $LC^2$ \cite{wein2007simulation} can perform better.


Deep networks have dominated medical imaging and machine vision in the past few years, proving powerful for many applications.  These networks automatically extract intermediate- and high-level representations of image structures that can be effectively
used for problem solving. In view of this, might it be that deep networks could automatically learn registration metrics that perform better than those developed by human designers? Recently, registration methods based on deep networks have been developed 
that use classifier technology to synthesize deep metrics for registration (DMR). Simonovsky \textit{et al.} \cite{simonovsky2016deep}  proposed an application specific deep metric based on Convolutional Neural Networks (CNNs). 
Although they showed superior performance compared to MI for deformable registration, they require well-registered training data,  i.e., the method is {\em fully supervised}. 
Balakrishnan \textit{et al.}
\cite{adrian2018AnUL} presented VoxelMorph, an {\em unsupervised}  approach, where registration is modeled as a parametric function using CNNs. During training, the model parameters are optimized by maximizing image similarity which does not require ground truth registration. However, the similarity metric in VoxelMorph is designed for intra-modality registration. In another work, Hu \textit{et al.} \cite{Hu2018WeaklysupervisedCN} described a {\em weakly supervised} approach where sparse corresponding landmarks are used to summarize the underlying dense deformation, with good results on the registration of prostate MR and ultrasound. 

We propose a new approach, Deep Information Theoretic Registration (DITR), that uses Iterated Maximum Likelihood (IML) to train effective application specific deep image metrics. We show that this approach is strongly related to MI, but on patches, not pixels. This alleviates one of the main limitations of the MI approach, namely the strong implicit independence assumption on pixels or voxels. In addition, we show that in our method, neither landmarks nor well-registered training data are required for learning the deep metric.  This is an important issue in some applications e.g., MRI and ultrasound, as it is not practical to perform simultaneous scanning, so that with soft tissues, accurate alignment of the  training data is unlikely.  

The remainder of the paper is structured as follows: Section 2 describes the relationship between Maximum Likelihood (ML) and 
information theoretic (IT) registration methods. In Section 3, we describe achieving ML registration using classification. Next we demonstrate that the need for accurately registered training data is relaxed by using IML 
with deep classifier technology. Finally, in Section 4 we evaluate our proposed method for multi-modality image registration, and explore data augmentation techniques that also help to obviate the need for accurately registered training data.

\section{Maximum Likelihood Registration}
Before introducing the relation between deep image metrics and MI registration, we provide a brief history of ML and MI based registration methods.  Given that ML is older than IT, it is perhaps interesting that ML registration appeared
after MI, in \cite{leventon1998multi}. We write ML registration of a fixed image $U$ and moving image $V$ as
\begin{equation*}
  \hat{\beta} = \argmax_\beta \ln p(U,V;\beta, \hat{\theta}) = \argmax_\beta \sum_i \ln p(u_i,v_i;\beta, \hat{\theta}) \PD
\end{equation*}
Here, $\beta$ are transformation parameters, and $\hat{\theta}$ are known modeling parameters,
that might have been previously estimated from training data
containing registered images.
We also assume that the images are collections of conditionally independent features,
$u_i,v_i$.
We model the joint distribution  on model features as
\begin{equation}
  p(u_i, v_i; \beta, \theta) \propto p_R(u_i, \TF{v_i}{\beta};\theta) \PD
\end{equation}

Here, $\TF{v}{\beta}$ is shorthand for $T(v,\beta)$ where $T$ is spatial transformation applied to the feature $v$.  $p_R(u_i,v_i;\theta)$ is intended as a joint
distribution on features {\em when registered}. 
In the present work, we suppress Jacobian effects, effectively
assuming that volume is approximately preserved.
The parameters of this
model could be estimated from training data consisting of registered
images.
Then,
\begin{equation}
  \hat{\beta} = \argmax_\beta \sum_i p_R(u_i, \TF{v_i}{\beta};\hat{\theta}) \PD
\end{equation}

This approach was used by Leventon \textit{et al.} \cite{leventon1998multi}.
In that work, the features were image pixels or voxels and the joint
distribution was categorical.   The model was estimated
by histogramming from pairs of registered images.
This need for pre-registered training data is a drawback in comparison to MI.

\subsection{Maximum Likelihood Registration, Unknown Parameters}

To proceed without training data, we could simultaneously optimize
the model parameters
when registering images.  If we view the model parameters as nuisance
parameters, this approach has been called {\em maximum profile likelihood} \;\cite{Cole2014MaximumLP}.
\begin{equation}
  \widehat{\beta, \theta} = \argmax_{\beta,\theta} \ln p(U,V;\beta, \theta) \PD
  \label{EqPFL}
\end{equation}
Here Bayesians may favor averaging over the nuisance parameters.
The two approaches are compared in 
\cite{severini1999relationship}.  In the context of registration, the marginalization
approach was described by Zollei \textit{et al.} \cite{zollei2007marginalized}.  We use the profile likelihood approach in the remainder of this paper.

\subsubsection{Special Case: Registration by Minimization of Joint Entropy}

We examine here the special case, as used in \cite{leventon1998multi}, where
the features are pixels, and the model is
categorical. Let the joint image intensities be discretized into bins that 
have a single index, calculated
by the function: $\INDEX(u_i, v_i)$.
Suppose $u_i, v_i$ are distributed according to a categorical distribution:
$\INDEX(u_i, v_i) | \theta \sim \CAT(\INDEX(u_i, v_i); \theta)$
where $\CAT(i;\theta) = \theta_i$,
$\theta_i \geq 0$ , and $\sum_i \theta_i = 1$. Then,
\begin{equation*}
\hat{\beta} = \argmax_\beta \max_\theta \sum_i \ln \theta_{\INDEX(u_i, \TF{v_i}{\beta})} = \argmax_\beta \max_\theta \left[\sum_j N_j(\beta) \ln \theta_j\right] \PD
\end{equation*}
where $i$ sums over all corresponding pairs of image intensities, 
$N_j(\beta)$ is the count of data items in bin $j$, 
and  $j$ sums over the bins.  
Note that the expression in brackets is the objective function for maximum
likelihood estimation of the parameters of the categorical distribution.
It is easy to show, using La Grange multipliers, that the expression
is maximized when $\hat{\theta_j}(\beta) = \frac{N_j(\beta)}{N}$
where $N \doteq \sum_i N_i(\beta$). Then (also dividing by $N$ inside the optimization),

\begin{equation*}
\hat{\beta} = \argmax_\beta \sum_j \frac{N_j(\beta)}{N}  \ln \frac{N_j(\beta)}{N} = \argmin_\beta \Ent{\CAT\left(\frac{N_j(\beta)}{N}\right)}\PD
\end{equation*}


Thus the optimization over $\beta$ has devolved exactly to minimization
of joint entropy.  This historically significant objective
function: ``adjust the registration so that the entropy of the 
joint histogram
is minimized'' \footnote{more accurately, the entropy of the categorical distribution corresponding to the normalized
joint histogram} 
\cite{collignon19953d} somewhat predates MI, they differ
by marginal entropy terms that may or may not be important in practice.  

\subsubsection{Optimization by Coordinate Ascent}
In cases where the inner optimization of Eq. \ref{EqPFL}
cannot be optimized in closed
form, we may use coordinate ascent by alternating
\begin{equation}
  \hat{\beta}^{n+1} = \argmax_{\beta}  \ln p(U, V,\beta, \hat{\theta^n})  
  \label{BetaOpt}
\end{equation}
and
\begin{equation}
  \hat{\theta}^{n+1} = \argmax_{\theta} \ln p(U, V,\hat{\beta^n}, \theta)  
  \label{ThetaOpt} \PD
\end{equation}

Eq. \ref{BetaOpt} amounts to (re-) estimating the transformation
parameters, given the (most recent) model parameter estimate.
Eq. \ref{ThetaOpt} re-estimates the model parameters from the
(re-) registered images.  We refer to this approach as ``Iterated
Maximum Likelihood (IML)'', it was used in Timoner's Phd thesis
\cite{timoner2003compact}.

\subsubsection{Information-Theoretic Interpretation}
Returning to the profile likelihood (Eq. \ref{EqPFL}),
\begin{equation}
  \widehat{\beta, \theta} = \argmax_{\beta,\theta} \ln p(U,V;\beta, \theta) = \argmax_{\beta,\theta}  \sum_i \ln p_R(u_i, \TF{v_i}{\beta};\theta) \PD
\end{equation}  

\noindent
Considering the data to be a sample, and 
using the asymptotic equivalence of sample average and expectation,
\begin{equation}
  \widehat{\beta, \theta} \approx \argmax_{\beta,\theta} \EV{p_D(u,\TF{v}{\beta})}{\ln p_R(u, \TF{v}{\beta};\theta)} \CM
\end{equation}
where $p_D(u,\TF{v}{\beta})$ is the (latent) true distribution of the features in the image (after the $v$ features have been transformed).
This can be re-written as
\begin{equation}
  \hat{\beta} \approx \argmin_{\beta}\left\{ \min_\theta \KLD{p_D(u,\TF{v}{\beta})}{p_R(u, \TF{v}{\beta};\theta)} + \Ent{p_D(u,\TF{v}{\beta})}\right\} \PD
\end{equation}

In some settings, if  $p_R$ has enough model capacity, the KL-divergence term may become unimportant, leaving 
only the joint entropy of the transformed
data. In this case, IML converges asymptotically to the minimum joint entropy of the feature data.  Similar relationships
among entropy and ML (on pixels or voxels) were described in \cite{zollei2003unified}.

\section{Maximum Likelihood Registration by Classifier}

In this section, we show how to use image classification to
generate an agreement metric
for solving registration problems.
Here a classifier is trained to distinguish between registered and unregistered patches. We 
use training data in form of $\{(u_i, v_i, z_i) ... \}$,
where $u_i$ and $v_i$ are features, or patches,
and $z_i \in \{0,1\}$.  
We construct
the data to contain a mix of well-registered pairs of patches,
labeled with $z_i = 1$, and
independently randomly  uniformly unregistered pairs, $z_i = 0$.  

We construct a discriminative classifier for this problem,
\begin{equation}
  p(z=1|u,v;\theta) \doteq \sigma(f(u,v,\theta)) \CM
  \label{EqDiscriminator} 
\end{equation}
where $\sigma(\cdot)$ is the sigmoid function, $f(\cdot) \in R$,
and $\theta$
represents the model parameters. $f(\cdot)$ can be any probabilistic classifier including a deep  network.
We use ML to train the conditional model (this is also called minimum cross entropy in the deep learning community).

\begin{equation*}
  \hat{\theta} = \argmax_\theta \sum_i \ln p(z_i|u_i,v_i; \theta) \PD
\end{equation*}

Next, we construct a joint distribution on registered patches
that is based on the classifier. From Bayes' rule, and noting that
$z_i$ does not depend on the parameters,

\begin{equation*}
  p(u_i,v_i|z_i;\theta)p(z_i) = p(z_i|u_i,v_i;\theta) p(u_i,v_i;\theta) \PD
\end{equation*}

Taking logs and subtracting over the two cases on the value of $z_i$,

\vspace{-1em}
\begin{multline*}
  \ln p(u_i,v_i|z_i= 1;\theta)  = \\[-0.35em] \ln p(u_i,v_i|z_i= 0;\theta)  + \ln \left(\frac{p(z_i=1|u_i,v_i;\theta)}{p(z_i=0|u_i,v_i;\theta)} \right)  -\ln\left(\frac{p(z_i=1)}{p(z_i=0)}\right) \PD
\end{multline*}

Note that the second term on the right  is the logit transform of 
$p(z_i=1|u_i,v_i;\theta)$, also that the logit transform is the inverse of $\sigma(\cdot)$.
Then using  Eq. \ref{EqDiscriminator}, and noting that the final term is a constant, 
\begin{equation}
  \ln p(u_i,v_i|z_i= 1;\theta) = \ln(p(u_i,v_i|z_i= 0;\theta)) + f(u_i,v_i;\theta) + C \PD
  \label{EqFoo}
\end{equation}

For the purpose of ML registration, 
we construct the joint distribution on patches conditioned on a transformation with parameters
$\beta$ as follows:
\begin{equation*}
  p(u_i,v_i;\beta, \hat{\theta}) \propto   p(u_i, \TF{v_i}{\beta}|z_i=1; \hat{\theta}) \PD
\end{equation*}

Taking log, and using Eq. \ref{EqFoo},

\begin{equation*}
  \ln   p(u_i,v_i;\beta, \hat{\theta}) = \ln p(u_, \TF{v_i}{\beta}|z_i=0)  + f(u_i, \TF{v_i}{\beta}) + C \PD
\end{equation*}

We argue that the first term on the right 
 above is constant in $\beta$,
 because, in the $z_i=0$ case, $u_i$ and $v_i$ are independent
 by assumption,
 we further assume that $p(v_i)$ is spatially stationary.  Then, we obtain ML registration as \\[-1.65em]

\begin{equation}
  \hat{\beta} = \argmax_\beta \sum_i \ln p(u_i, v_i;\beta,\hat{\theta})  \approx \argmax_\beta \sum_i f(u_i, \TF{v_i}{\beta},\hat{\theta}) \PD
\label{DeepMetricRegistration}
\end{equation}
This ML objective function is simply the sum of the pre-sigmoid network
responses  over the corresponding patches in the pair of images being registered;
it is our approach to DMR.

\subsection{Unsupervised Registration by Iterated Maximum Likelihood}

While DMR has proven useful, it is assumed that well-registered training data
is needed.  To relax this requirement, we use the IML
approach of Eqns. \ref{BetaOpt} and \ref{ThetaOpt};
in this context the training is over features in collections
of images,
\begin{equation}
  \hat{\beta}^{n+1} = \argmax_{\beta}  \sum_i \ln p(u_i, v_i; \beta, \theta^n)  \approx \argmax_\beta \sum_i f(u_i, \TF{v_i}{\beta},\hat{\theta^n})
  \label{EqBetaDeep}
\vspace{-1em}
\end{equation}
and,
\begin{equation}
  \hat{\theta}^{n+1} = \argmax_{\theta} \sum_i \ln p(u_i, v_i,\hat{\beta^n}, \theta)  \approx \argmax_\theta \sum_i \ln p(z_i|u_,\TF{v_i}{\hat{\beta^n}};\theta) \PD
  \label{EqThetaDeep}
\end{equation}
Eq. \ref{EqBetaDeep} amounts to estimating the transformation parameters
of a collection of pairs of images using
the deep metric with known model parameters, $\theta^n$.
Eq. \ref{EqThetaDeep} corresponds to retraining the network using patches that are offset by the most
recently estimated transformation parameters.
The iteration is started with Eq. \ref{EqThetaDeep} on the 
original roughly registered training data. Subsequently, 
the method alternates between re-aligning the data
and re-estimating the deep network parameters.
In the experimental 
portion of the paper, three iterations are applied in training.
We envision that this iterative training needs to happen only once per application type.
After training, the model parameters may be fixed and used for subsequent
registrations using Eq. \ref{DeepMetricRegistration}.


\section{Experimental Evaluation}
We perform several experiments to study the 
effectiveness of IML approach for several image registration tasks.
We introduce ``dithering''
as a tool to help demonstrate that the proposed approach does
not require perfectly aligned training data to learn
an accurate deep metric.
To generate  unregistered datasets, we perturb the moving images with a
random transformation. We select 100 landmarks in the image space to calculate and report mean Fiducial Registration Error (FRE)
following registration.
We compare our method to MI, a metric used often for multi-modality image registration, as well as multi-scale MI registration (with 75 histogram bins). 
In all of our experiments, we use Powell's method \cite{powel}
for optimization of the transformation parameters (Eq. \ref{EqBetaDeep}) with the
learned deep metric as a cost function.

\subsection{Data}
We carried out experiments using the IXI Brain
Development Dataset \cite{BrainDev33:online} which contains
aligned T1-T2 image pairs from healthy subjects.
In our experiments, we use 60 subjects for training and another 60 subjects for validation.
All images are resampled to $1{\times}1{\times}1~mm$, and their intensity is normalized
between the range of $\left[0,1\right]$. We crop 3D patches, $u_i,v_i$, of size $17{\times}17{\times}17$
containing a mix of registered pairs of patches with $z_i=1$ (cropped from the same location in image space), and randomly uniformly unregistered pairs with $z_i=0$. Overall, 1 million patches
are generated for each experiment. 

\textbf{Dithering} 
Training a deep metric on unregistered dataset can cause bias in
the response function depending on the distribution of the
mis-registration in the data. For example, if the moving images were shifted in the $x$ direction, the response function will have a peak that is shifted accordingly (see Fig. 3a). Data augmentation with rotation and flipping can help reduce this bias at a cost of introducing additional variance and peaks (modes) in the
response function. A smooth, single peak response function is preferred for effective optimization and learning of the transformation parameters. Traditionally, this could be remedied via image smoothing, however we show below that with DMR, smoothing is ineffective. Therefore, we propose ``dithering'' as an effective alternative approach to merge the multiple modes of 
the response function. More specifically, we deliberately introduce noise by applying
3D Gaussian distributed random displacements to the moving image
prior to cropping patches $v(x_i) \rightarrow v(x_i+d)$ 
where $d \sim N(0,\sigma^2I)$ and $\sigma$ is the standard deviation
of the dither. 

\vspace{-2mm}
\subsection{Network Architecture and Training}

For learning a similarity metric for image registration, we use
3D CNNs and train the networks as discriminative
classifiers to distinguish between registered and unregistered image patches
in our experiments. 

The architecture of our network is inspired by the 2-channel
network of Zagouruko \textit{et al.} \cite{Zagoruyko2015LearningTC} where patches from the fixed and moving images are the input channels of the CNN. The network has a 5-layer architecture consisting of strided 3D convolutions of size
$3{\times}3{\times}3$ and ReLU activation functions followed by
an average pooling layer and a sigmoid.

\textbf{CNN Training and Registration} We train our model by minimizing cross-entropy loss. During training, a learning rate of $5{\times}10^{-5}$, batch size of $256$ and $\ell_2$-regularization (weight decay) of $0.005$ is used to optimize the network. Following training, we use the sum of the pre-sigmoid
network responses over the patches in the pair of images being registered as our cost
function for optimization to update the transformation parameters. The process of training and transformation update may be iterated if necessary to improve registration of the training data.

\subsection{Experiments}
\textbf{Experiment 1: Rigid and Affine Registration} In this experiment,
we demonstrate the effectiveness of IML approach in learning a deep metric
from roughly registered pairs of images. In addition, we evaluate the contribution of dithering
to registration by comparing the performance of IML with and without dithering.
First, we perform a rigid registration experiment where we perturb
the moving images by applying a random rigid transformation with
parameters sampled from a 3D uniform distribution of
$\mathcal{U}_{t}\{1,25\}~mm$ and $\mathcal{U}_{\theta}\{0.01,0.15\}~rad$
for translation and rotation, respectively.
We propose 3 iterations of IML as $\text{IML}_1(4,100)\rightarrow
\text{IML}_2(2,15) \rightarrow \text{IML}_3(0,0)$ where
$\text{IML}(l,\sigma^2)$ represents training a model with images that
are donwnsampled by factor of $l$ and
moving images dithered with a variance of $\sigma^2$. Downsampling
is needed because large initial mis-registrations cannot be captured 
by the limited patch size. 

Furthermore, we experiment with affine registration to show the capability
of IML in a more general registration problem.
We perturb the moving images by applying
a random transformation with parameters sampled from
$\mathcal{U}_{t}\{1,25\}$, $\mathcal{U}_{\theta}\{0.01,0.15\}$,
$\mathcal{U}_{s}\{0.95,1.05\}$ and $\mathcal{U}_{sh}\{-0.01,0.01\}$ 
for translation, rotation, scale and shear, respectively.
We follow the 3 stage IML model proposed earlier to learn the deep metrics.
We also characterize the deep metric learned from the roughly registered data
as a function of translation for a test data (before perturbation),
following each iteration of IML, to illustrate the nature of the
objective functions.

\textbf{Experiment 2: Effect of Dithering on Response Function}
In order to experimentally verify the value of dithering as a way to
merge the modes of the response function, we compare three methods
of training on data where moving images are systematically
shifted $10~mm$ in the \textit{x} direction.  This is meant to
simulate a situation that could occur if there were a consistent difference in
setup between non-simultaneous scanning of different image modalities.
These three methods are: (a) training a model on the shifted data,
without any augmentation and dithering (b) training a model on the same data,
with augmentation, without dithering, and (c) training
a model on the same data, with augmentation and dithering.
In another experiment,
to compare dithering and smoothing for broadening the response function, we perform an experiment in which we learn a deep metric on the
dithered and smoothed data separately.

\textbf{Experiment 3: Edge Registration} We test our proposed
IML approach in a more difficult multi-modality situation.
More specifically, we experiment with registration of 
edge maps of the T1 images (extracted by the Canny edge detector) to
intensities in T2 images.
We apply the edge detector to the
same data that was used in the rigid registration experiment, and follow 3 iterations of IML to learn the deep metric,
$\text{IML}_1(2,20)\rightarrow \text{IML}_2(2,10) \rightarrow \text{IML}_3(0,0)$.

\section{Results and Discussion}
\begin{figure}[t]
\centering
{\includegraphics[width=\linewidth]{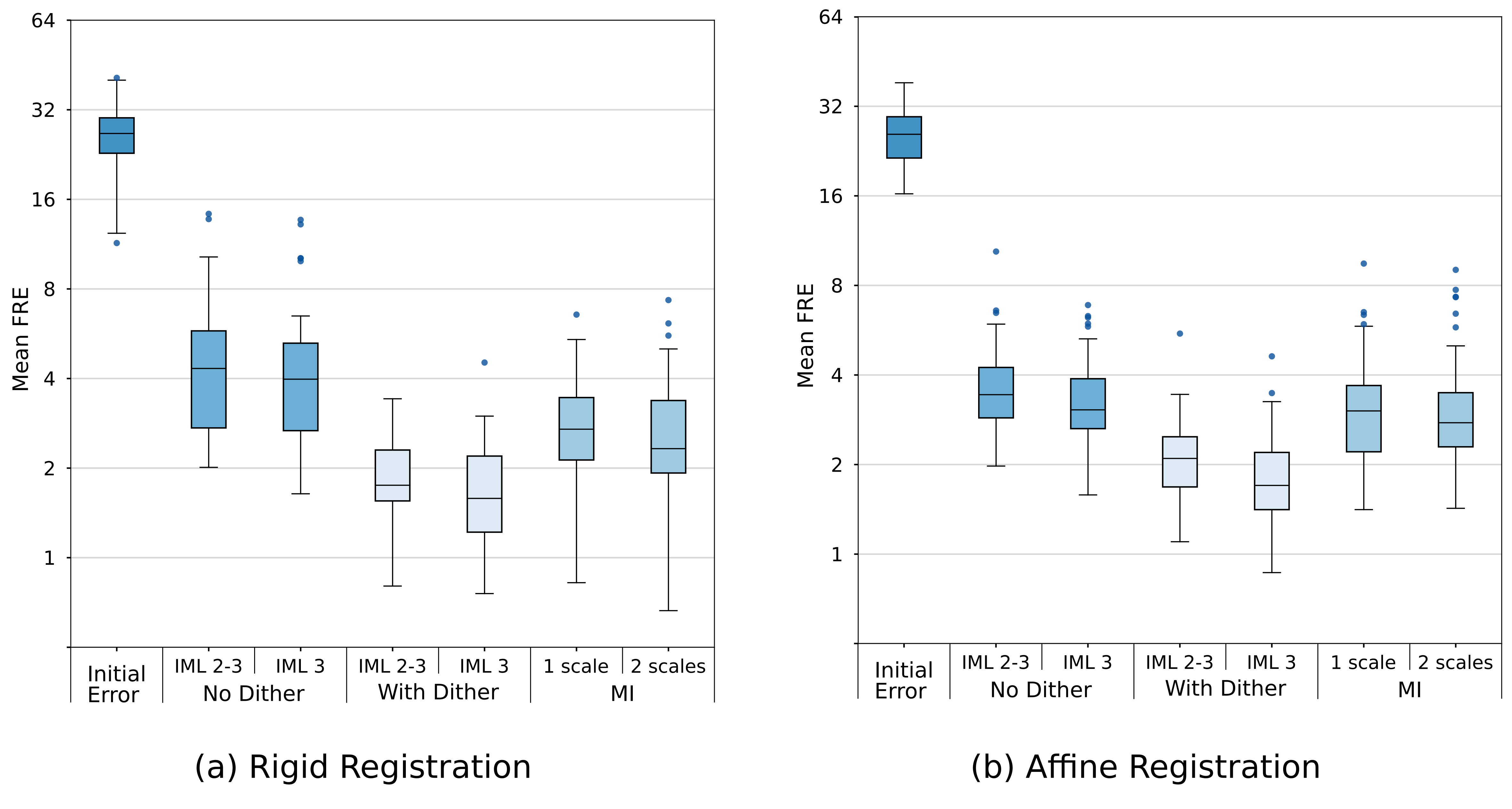}}
\caption{Box plots of mean FRE for rigid (a) and affine (b) registration between T1 and T2 images.}
\label{fig:rigid}
\end{figure}

Fig. \ref{fig:rigid} shows box plots of mean FRE for different experiments performed for rigid and affine registration. Each box represents the interquartile range, and the horizontal line is the median of the distribution of mean FRE. As seen in Fig. \ref{fig:rigid}a, IML with dithering performs statistically significantly better than IML without dithering (two-sided t-test {$p<0.001$}), and MI ($p<0.001$). Moreover, \mbox{Fig. \ref{fig:rigid}b} demonstrates the effectiveness
of IML with dithering for affine registration compared to MI ($p<0.001$). For both registration tasks, IML without dithering improves the initial registration error to some extent which further demonstrates the added value of dithering for learning deep metrics from unregistered datasets.

\begin{figure}[h!]
\centering
{\includegraphics[width=\linewidth]{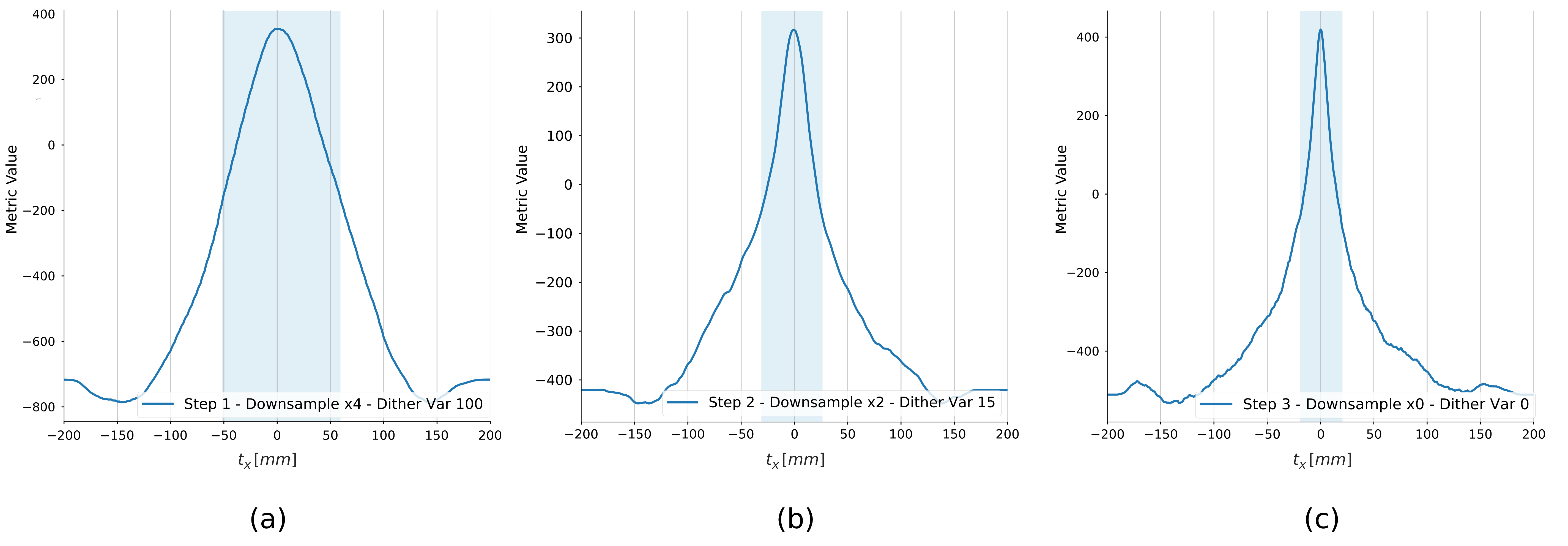}}
\caption{Response functions for each iteration of IML for rigid registration plotted as a function of translation for a pair of
registered fixed and moving images. Shading illustrates full-width half max.}
\label{fig:activations}
\end{figure}

Fig. \ref{fig:activations} delineates the deep metric on pairs of images as a function of translation in each iteration of IML.
We observe that from the first iteration, IML(4,100), to the last iteration, the response function is sharper and hence the accuracy of the image registration improves. This is likely due to the increased level of alignment of the training data as the iterations proceed.

\begin{figure}[t]
\centering
\vspace{-1em}
{\includegraphics[width=0.9\linewidth]{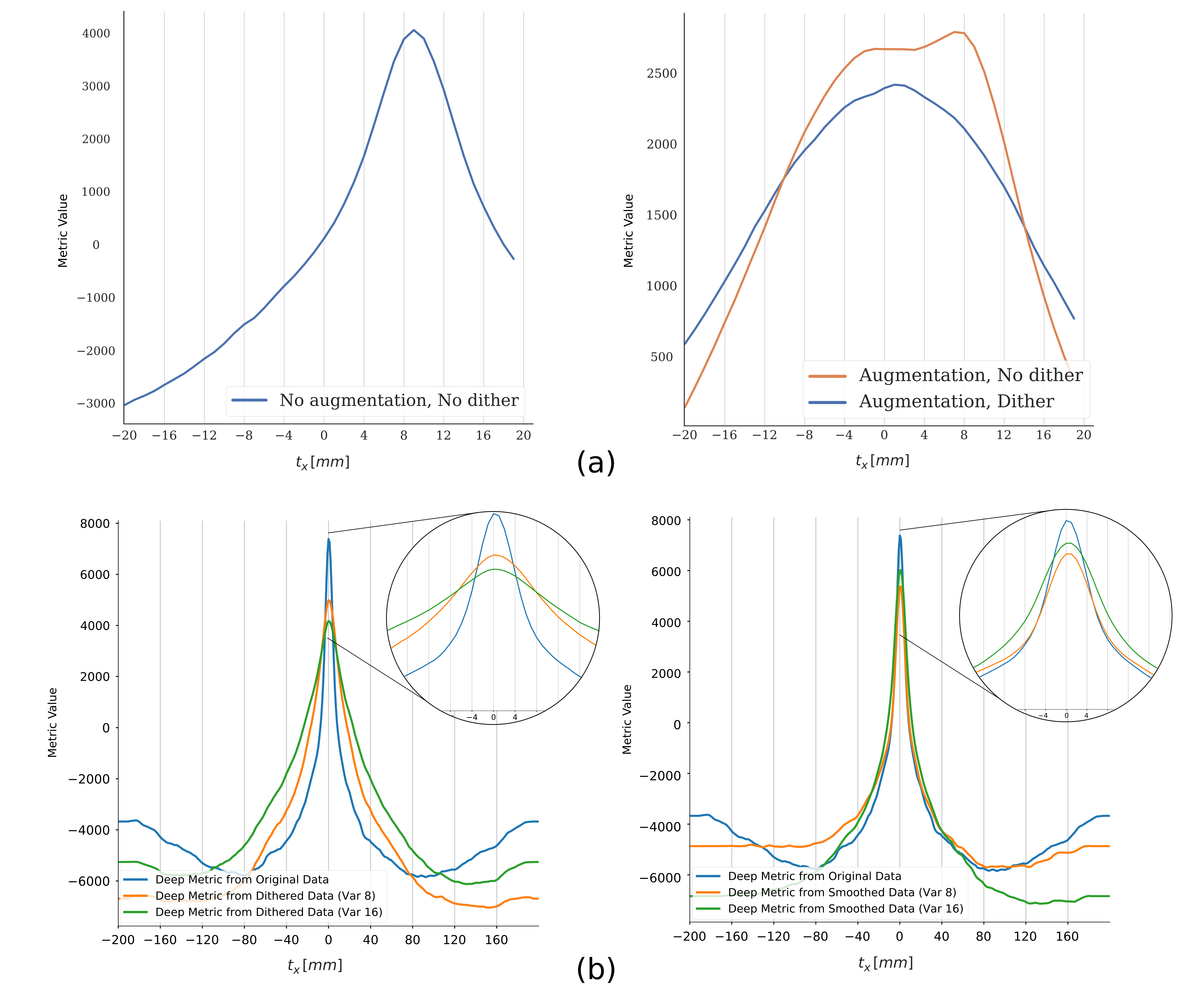}}
\caption{(a) Characteristics of the deep metric learned from different training data. (b) Impact of dithering (left) and
smoothing (right) on the deep metric response function.}
\label{fig:dither}
\end{figure}

Fig. \ref{fig:dither} shows the results of Experiment 2 for exploring the effect of dithering on learning deep image metrics.  In Fig. \ref{fig:dither}a, it is clear that there is a bias in the response function due to the systematic shift in the training data. Fig. \ref{fig:dither}a also shows the effect of augmentation by rotation and flipping in reducing the bias. Moreover, we can see that by applying dithering to the moving image before cropping the patches, we are able to merge the peaks (modes) in  the deep metric. We evaluate the impact of dithering versus smoothing on the broadness of the deep metric in Fig. \ref{fig:dither}b.
It is interesting to note that smoothing has not significantly broadened the response function. We believe that deep networks are capable of effectively learning the correspondence between the smoothed patches; therefore, they can generate a sharp response similar to the response function of the deep metric learned from the original data. We see on the left that dithering is more effective at broadening the response. 

\begin{figure}[t]
\centering
\vspace{0cm}
{\includegraphics[width=\linewidth]{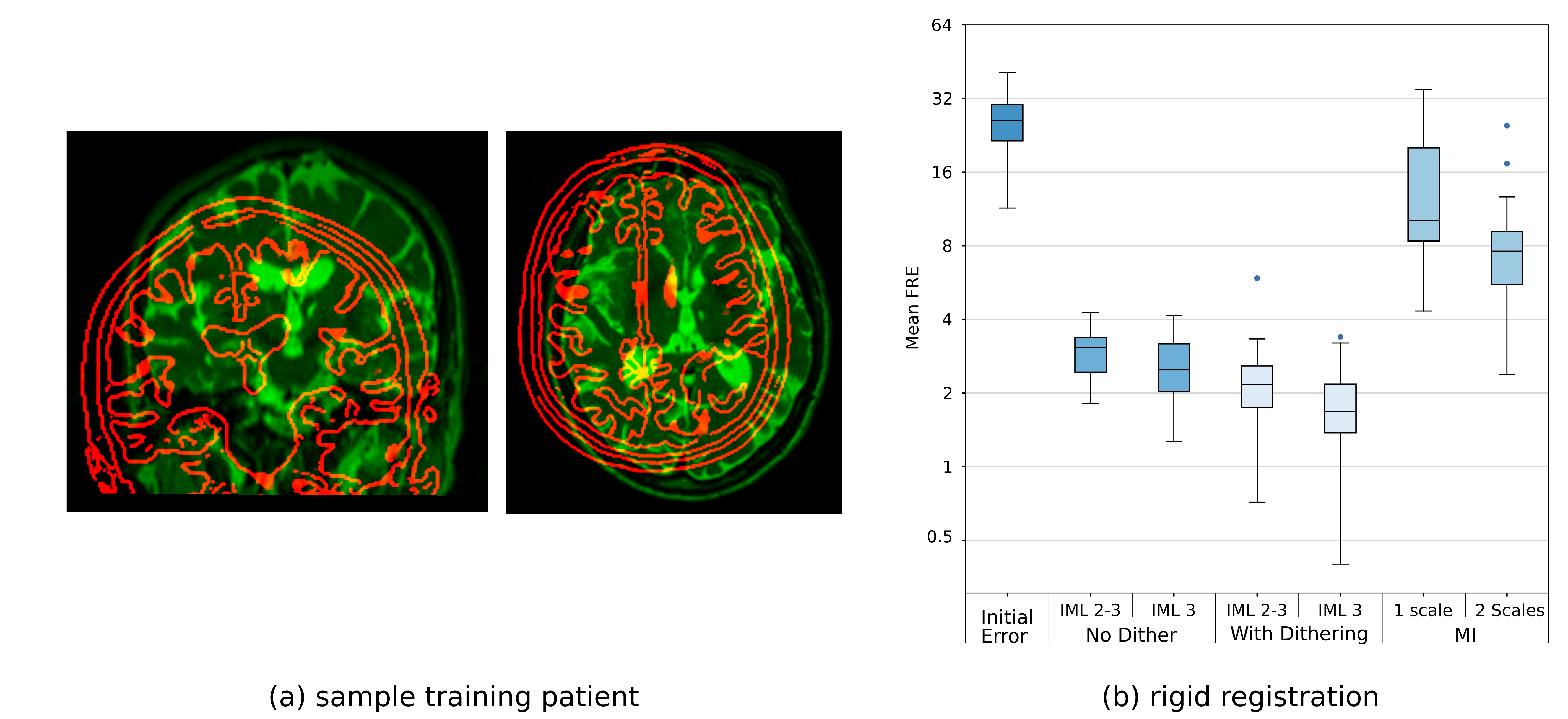}}
\caption{(a) An example case from the edge-to-image registration experiment (b) mean FRE achieved by different methods.}
\label{fig:canny}
\end{figure}

Fig. \ref{fig:canny}a depicts sample image pairs from the training dataset for Experiment 3, Edge Registration, with the initial  mis-registration. Figure \ref{fig:canny}.b shows the mean FRE using different methods. The figure clearly demonstrates the superior performance of DITR compared to MI. 

In the experiments presented, several iterations of IML are needed to learn the best deep metrics, but for registering the test data we  only need the final trained network. We believe this is possible due to the broad capture range of the learned deep metric (Fig. \ref{fig:activations}). 


\section{Conclusion}
\vspace{-2.5mm}
We presented, for the first time, an information theoretical (IT) foundation for iterated maximum likelihood (IML) registration with deep image metrics, DITR. We further showed that IML with classifier-based metrics is strongly related to mutual information
on  patches (not pixels). 

We expect that DITR will enable new solutions for  applications where the standard MI assumption of pixel- or voxel-wise independence is limiting. We demonstrated the effectiveness of our proposed method in rigid and affine registration for multi-modal data. In all experiments, DITR outperformed standard MI statistically significantly. On the edge-to-image registration experiment, MI effectively failed, but DITR  successfully registered the images. 

Our work focused on  the analysis of registration objective functions rather than transformation modeling and optimization methods; we expect the technology to be effective in more general settings, as it is a generalization of similar DMR methods that have been shown to be successful for inter-subject deformable registration
\cite{simonovsky2016deep}.

\section{Acknowledgements}
Research reported in this publication was supported by Natural Sciences and Engineering Research Council (NSERC) of Canada, the Canadian Institutes of Health Research (CIHR), Ontario Trillium Scholarship, NIH Grant No. P41EB015898, and NIH NIBIB Grant No. P41EB015902 Neuroimage Analysis Center.

\bibliographystyle{unsrt}
\bibliography{references}

\end{document}